\documentclass[lettersize,journal]{IEEEtran}
\usepackage{amsmath,amsfonts}
\usepackage{array}
\usepackage{textcomp}
\usepackage{stfloats}
\usepackage{url}
\usepackage{verbatim}
\usepackage{graphicx}
\usepackage{cite}
\usepackage{mathrsfs}
\usepackage[colorlinks,linkcolor=blue,anchorcolor=blue,citecolor=blue]{hyperref}
\usepackage{babel}
\usepackage{amsmath}
\usepackage{amssymb}
\usepackage{mathtools}
\usepackage{multirow}
\usepackage{booktabs}
\usepackage{tabularx}
\usepackage{pgfplots}
\usepackage{xcolor}
\usepackage{subcaption}
\usepackage[ruled,linesnumbered]{algorithm2e}
\usetikzlibrary{patterns}
\usepackage{orcidlink}

\def\BibTeX{{\rm B\kern-.05em{\sc i\kern-.025em b}\kern-.08em
    T\kern-.1667em\lower.7ex\hbox{E}\kern-.125emX}}
\usepackage{balance}

\begin{document}

\title{Knowledge Graphs and Pre-trained Language Models enhanced Representation Learning for Conversational Recommender Systems}

\author{
    Zhangchi~Qiu~\orcidlink{0000-0001-8763-4070},
    Ye~Tao~\orcidlink{0000-0002-0996-9833},
    Shirui~Pan~\orcidlink{0000-0003-0794-527X},~\IEEEmembership{Senior Member,~IEEE} \\
    and~Alan~Wee-Chung~Liew$^{*}$~\orcidlink{0000-0001-6718-7584},~\IEEEmembership{Senior Member,~IEEE,}
    \thanks{This research was supported in part by the Australian Research Council (ARC) under grants FT210100097 and DP240101547, and the CSIRO – National Science Foundation (US) AI Research Collaboration Program. Zhangchi Qiu is supported by the Griffith University Postgraduate Research Scholarship. Manuscript received xx; revised xxx; accepted xxx. Date of publication xxx; date of current version xxx.
        \textit{(Corresponding author: Alan~Wee-Chung~Liew.)}}
    \thanks{Zhangchi Qiu, Ye~Tao, Shirui~Pan, and Alan~Wee-Chung~Liew are with the School of Information and Communication Technology, Griffith University, Gold Coast, Queensland 4222, Australia (e-mail: \{zhangchi.qiu, ye.tao2\}@griffithuni.edu.au; \{s.pan, a.liew\}@griffith.edu.au).}
}

\markboth{Journal of \LaTeX\ Class Files,~Vol.~14, No.~8, August~2021}%
{Shell \MakeLowercase{\textit{et al.}}: A Sample Article Using IEEEtran.cls for IEEE Journals}

\IEEEpubid{0000--0000/00\$00.00~\copyright~2021 IEEE}

\maketitle

\begin{abstract}
    Conversational recommender systems (CRS) utilize natural language interactions and dialogue history to infer user preferences and provide accurate recommendations.
    Due to the limited conversation context and background knowledge, existing CRSs rely on external sources such as knowledge graphs to enrich the context and model entities based on their inter-relations.
    However, these methods ignore the rich intrinsic information within entities.
    To address this, we introduce the Knowledge-Enhanced Entity Representation Learning (KERL) framework, which leverages both the knowledge graph and a pre-trained language model to improve the semantic understanding of entities for CRS.
    In our KERL framework, entity textual descriptions are encoded via a pre-trained language model, while a knowledge graph helps reinforce the representation of these entities.
    We also employ positional encoding to effectively capture the temporal information of entities in a conversation.
    The enhanced entity representation is then used to develop a recommender component that fuses both entity and contextual representations for more informed recommendations, as well as a dialogue component that generates informative entity-related information in the response text.
    A high-quality knowledge graph with aligned entity descriptions is constructed to facilitate our study, namely the Wiki Movie Knowledge Graph (WikiMKG).
    The experimental results show that KERL achieves state-of-the-art results in both recommendation and response generation tasks. Our code is publicly available at the link: \url{https://github.com/icedpanda/KERL}.
\end{abstract}


\IEEEpeerreviewmaketitle

\section{Introduction}
\IEEEPARstart{O}{ver} the years, recommendation systems have become increasingly popular due to the growing demand for personalized recommendations.
Traditional recommendation techniques, such as collaborative filtering~\cite{sarwar_item_based_2001, schafer_collaborative_2007, das_abhinandan_google_2007} and content-based filtering~\cite{pazzani_content_based_2007, Lops_content_rec_sota_2011}, depend solely on users' historical interactions to generate suggestions.
However, these methods encounter several drawbacks. They may yield recommendations misaligned with the user's current interests, often suggesting items that are similar to those previously interacted with. Furthermore, they are not adept at capturing sudden changes in user preferences, rendering them less responsive to the user's evolving interests.

\IEEEpubidadjcol

These drawbacks motivated the development of conversational recommender systems (CRSs)~\cite{Christakopoulou2016TowardsCR, serban_hierarchical_2017, chen-etal-2019-towards, zhou_improving_2020, lu_revcore_2021}, which seek to address the limitations of traditional recommendation systems by employing natural language processing (NLP) techniques to engage users in multi-turn dialogues, allowing CRS to elicit their preferences, and provide personalized recommendations that are better aligned with their current situations, along with accompanying explanations. As an example shown in Table~\ref{tab:crs}, a CRS aims to recommend a movie suitable for the user. To accomplish this, the CRS initially offers a recommendation based on the user's expressed preferences. As the user provides feedback, the CRS refines its understanding of the user's current interests and adjusts its recommendations accordingly, ensuring a more personalized and relevant suggestion.

\begin{table}[tb]
  \small
  \centering
  \caption{An illustrative example of a chatbot-user conversation on movie recommendations, with items (movies) in italic blue font and entities (e.g., movie genres) in italic red font.}
  \begin{tabularx}{\linewidth}{lX}
    \toprule
    \textbf{User}     & Hi, I really like \textit{\textcolor{red}{\textbf{comedies}}}. Can you recommend one? Something like \textit{\textcolor{blue}{\textbf{Grease (1978)}}}? \\ \midrule
    \textbf{Chatbot}  & Have you seen \textit{\textcolor{blue}{\textbf{Step Brothers (2008)}}}? It is really funny.                                                             \\ \midrule
    \textbf{User}:    & Yes, I have seen it. It's great! Maybe a \textit{\textcolor{red}{\textbf{musical}}} too, like \textit{\textcolor{blue}{\textbf{Chicago (2002)}}}?       \\ \midrule
    \textbf{Chatbot}: & If you like that type, you'd probably enjoy \textit{\textcolor{blue}{\textbf{Wedding Crashers (2005)}}}. It's similar.                                  \\ \midrule
    \textbf{User}:    & I haven't seen that one yet. That sounds perfect! Thanks!                                                                                               \\ \bottomrule
  \end{tabularx}
  \label{tab:crs}
\end{table}
To develop an effective CRS, existing studies~\cite{chen-etal-2019-towards, zhou_improving_2020, zhang-etal-2022-toward} have explored the integration of external data sources, such as knowledge graphs (KGs) and relevant reviews~\cite{lu_revcore_2021}, to supplement the limited contextual information in dialogues and backgrounds.
These systems extract entities from the conversational history and search for relevant candidate items in the knowledge graph to make recommendations.

\begin{figure}[t]
  \centering
  \includegraphics[width=\columnwidth]{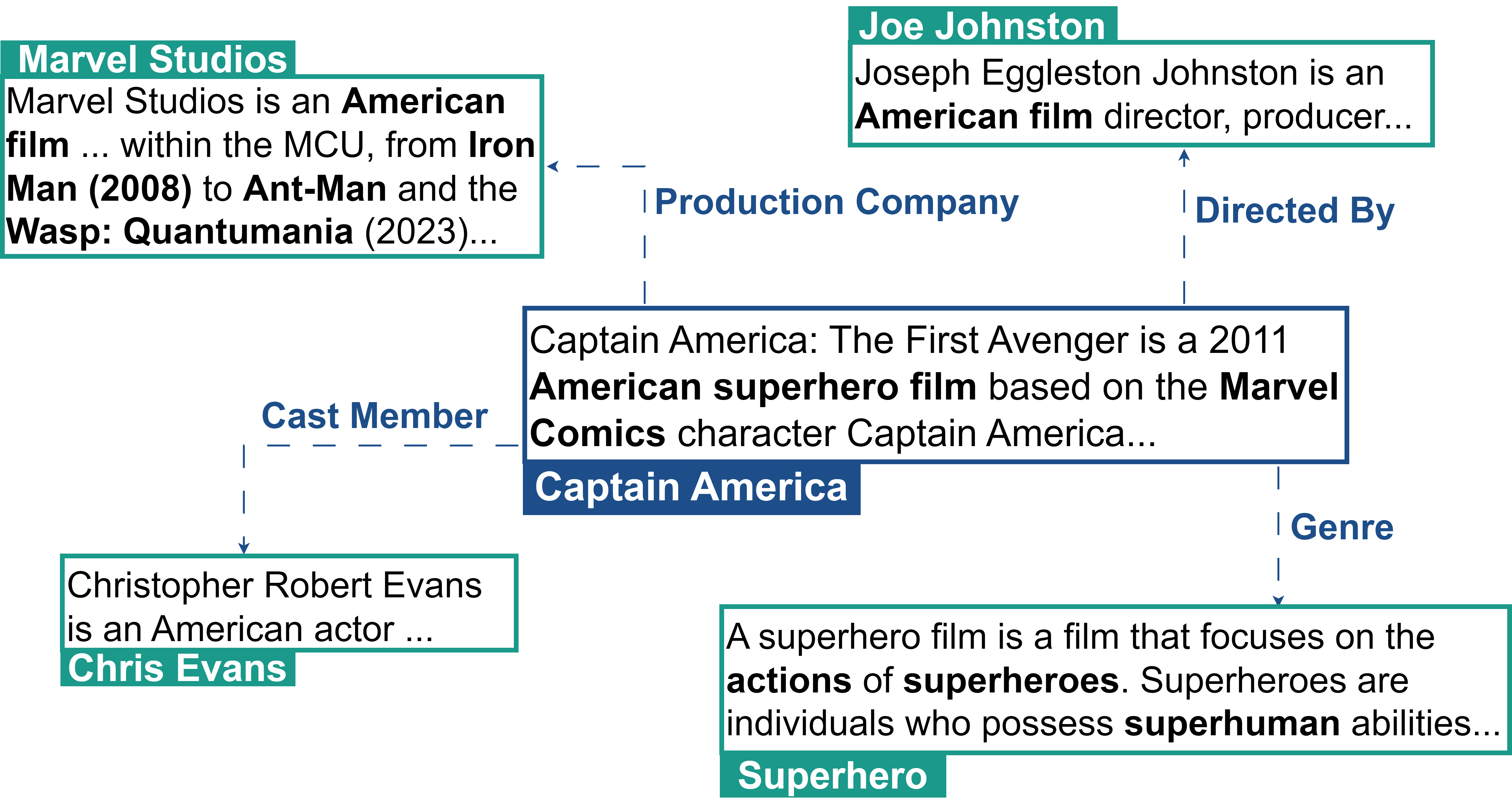}
  \caption{Example of a KG that incorporates entity descriptions. The figure suggests that descriptions contain rich information and can help improve the semantic understanding of entities}
  \label{fig:kg}
\end{figure}

Despite the advances made by these studies, three main challenges remain to be addressed:
\begin{itemize}
  \item \textbf{Challenge 1}: Current systems tend to model entities solely on the basis of their relationships within the KG, failing to exploit the valuable information contained in textual descriptions, as shown in Figure~\ref{fig:kg}.
    Using only the topological information from a knowledge graph may not capture the nuanced meanings and knowledge associated with the entities.
       
  \item \textbf{Challenge 2}: Previous research has overlooked the importance of the sequence order of entities in the context by treating them as a set rather than a sequence. This approach is contrary to the goal of CRS, as the order in which entities appear in the conversation can significantly impact the conversation's topic and result in different conversational trajectories. Ignoring the sequence information can lead to recommendations that do not accurately reflect the user's current interests and context, resulting in suboptimal performance.
  \item \textbf{Challenge 3}: The responses generated by these systems have been observed to lack diversity and comprehensive information, and often fail to provide a description of the items being recommended.
\end{itemize}

To address these issues, we propose a novel framework that leverages KGs and pre-trained language models (PLMs) for enhanced representation learning in CRSs.
This framework, which we refer to as KERL, stands for \textbf{K}nowledge-enhanced \textbf{E}ntity \textbf{R}epresentation \textbf{L}earning framework.
Firstly, we use a PLM to directly encode the textual descriptions of entities into embeddings, which are then used in combination with graph neural networks to learn entity embeddings that incorporate both topological and textual information. This results in a more comprehensive representation of the entities and their relationships (challenge 1).
Secondly, we adopt a positional encoding inspired by the Transformer~\cite{vaswani_attention_2017} to account for the sequence order of entities in the context history (challenge 2), along with a conversational history encoder to capture contextual information in the conversation. By combining these two components, we are able to gain a better understanding of the user's preferences from both entity and contextual perspectives.
This leads to more informed and tailored recommendations that better align with the user's current interests and context.
However, as these two user preference embeddings are from two different embedding spaces, we employ the contrastive learning \cite{chen_2020_simple} method to bring together the same users with different perspectives, such as entity-level user preferences and contextual-level user preferences, while simultaneously distancing irrelevant ones.
Lastly, we integrate the knowledge-enhanced entity representation with the pre-trained BART~\cite{lewis-etal-2020-bart} model as our dialogue generation module.
This allows us to leverage the capability of the BART model while also incorporating entity knowledge to generate more diverse and informative responses (challenge 3), providing a more comprehensive and engaging experience for the user.

The contributions can be summarized as follows:
(1) We construct a movie knowledge graph (WikiMKG) with entity description information.
(2) We use a knowledge-enhanced entity representation learning approach to enrich the representation of entities that captures both topological and textual information.
(3) We utilize positional encoding to accurately capture the order of appearance of entities in a conversation. This allows for a more precise understanding of the user's current preferences.
(4) We adopt a contrastive learning scheme to bridge the gap between entity-level user preferences and contextual-level user preferences.
(5) We integrate entity descriptions and a pre-trained BART model to improve the system's ability to compensate for limited contextual information and enable the generation of informative responses.

\section{Related Work}\label{sec:related work}

\subsection{Conversational Recommender System}
With the rapid development of dialogue systems~\cite{serban-2016-build, chen-etal-2017-deep, Zhang2019DIALOGPTL, lian_pirnet_2022, wang_target-driven_2023}, there has been growing interest in utilizing interactive conversations to better understand users' dynamic intent and preferences.
This has led to the rapidly expanding area of conversational recommender systems \cite{Christakopoulou2016TowardsCR, ConversationalRecommenderSystem_2018a, li_towards-deep_2018}, which aim to provide personalized recommendations to users through natural language interactions.

In the realm of CRS, one approach involves the use of predefined actions, such as item attributes and intent slots~\cite{Christakopoulou2016TowardsCR, ConversationalRecommenderSystem_2018a, interactive_path_lei, zhang_conversation-based_2023-1}, for interaction with users.
This category of CRS primarily focuses on efficiently completing the recommendation task within a limited number of conversational turns. To achieve this objective, they have adopted reinforcement learning~\cite{ConversationalRecommenderSystem_2018a, EstimationActionReflectionDeepInteraction_2020a, InteractiveRecommenderSystem_2020, zhang_conversation-based_2023-1}, multi-armed bandit~\cite{Christakopoulou2016TowardsCR} to help the system in finding the optimal interaction strategy. However, these methods still struggle to generate human-like conversations, which is a crucial aspect of a more engaging and personalized CRS.

Another category of CRS focuses on generating both accurate recommendations and human-like responses, by incorporating a generation-based dialogue component in their design. Li et al.~\cite{li_towards-deep_2018} proposed a baseline~HRED-based~\cite{serban_hierarchical_2017} model and released the CRS dataset in a movie recommendation scenario. However, the limited contextual information in dialogues presents a challenge in accurately capturing user preferences.
To address this issue, existing studies introduce the entity-oriented knowledge graph~\cite{chen-etal-2019-towards, zhang-etal-2022-toward}, the word-oriented knowledge graph~\cite{zhou_improving_2020}, and review information~\cite{lu_revcore_2021}. 
This information is also used in text generation to provide knowledge-aware responses. Although these integrations have enriched the system's knowledge, the challenge of effectively fusing this information into the recommendation and generation process still remains. Therefore, Zhou et al.~\cite{zhou_c-crs_2022} proposed the contrastive learning approach to better fuse this external information to enhance the performance of the system. Additionally, instead of modeling entity representation with KG, Yang et al.~\cite{yang-etal-2022-improving} constructed entity metadata into text sentences to reflect the semantic representation of items. However, such an approach lacks the capability to capture multi-hop information. 

Inspired by the success of PLMs, Wang et al.~\cite{wang-etal-2022-recindial} combined DialogGPT~\cite{Zhang2019DIALOGPTL} with an entity-oriented KG to seamlessly integrate recommendation into dialogue generation using a vocabulary pointer. 
Furthermore, Wang et al.~\cite{wang_towards_2022} introduced the knowledge-enhanced prompt learning approach based on a fixed DialogGPT~\cite{Zhang2019DIALOGPTL} to perform both recommendation and conversation tasks.
These studies do not exploit the information present in the textual description of entities and their sequence order in the dialogue. In contrast, our proposed KERL incorporates a PLM for encoding entity descriptions and uses positional encoding to consider sequence order, leading to a more comprehensive understanding of entities and conversations.

\subsection{Knowledge Graph Embedding}

Knowledge graph embedding (KGE) techniques have evolved to map entities and relations into low-dimensional vector spaces, extending beyond simple structural information to include rich semantic contexts. These embeddings are vital for tasks such as graph completion \cite{xie_representation_2016, wang_logic_2019, li_adaptive_2023}, question answering~\cite{calders_open_2014, huang_knowledge_2019}, and recommendation~\cite{zhang_collaborative_2016, wang_dkn_2018}. Conventional KGE methods such as TransE~\cite{bordes_2013_translating}, RotatE~\cite{sun2018rotate} and DistMult~\cite{yang_embedding_2015} focus on KGs' structural aspects, falling into translation-based or semantic matching categories based on their unique scoring functions~\cite{wang_knowledge_2017, ji_survey_2022}. 
Recent research capitalizes on the advancements in NLP to encode rich textual information of entities and relations. DKRL~\cite{xie_representation_2016} were early adopters, encoding entity descriptions using convolutional neural networks. Pretrain-KGE~\cite{zhang_pretrain-kge_2020} further advanced this approach by employing BERT~\cite{devlin_bert_2019} as an encoder and initializes additional learnable knowledge embeddings, then discarding the PLM after fine-tuning for efficiency. Subsequent developments, including KEPLER~\cite{wang_kepler_2021} and JAKET~\cite{yu_jaket_2022}, have utilized PLMs to encode textual descriptions as entity embeddings. These methods optimize both knowledge embedding objectives and masked language modeling tasks. Additionally, LMKE~\cite{wang_language_2022} introduced a contrastive learning method, which significantly improves the learning of embeddings generated by PLMs for KGE tasks. In comparison to these existing methods, our work enhances conversational recommender systems by integrating a PLM with a KG to produce enriched knowledge embeddings. We then align these embeddings with user preferences based on the conversation history. This method effectively tailors the recommendation and generation tasks in CRS.

\begin{figure*}[t]
  \centering
  \includegraphics[width=2\columnwidth]{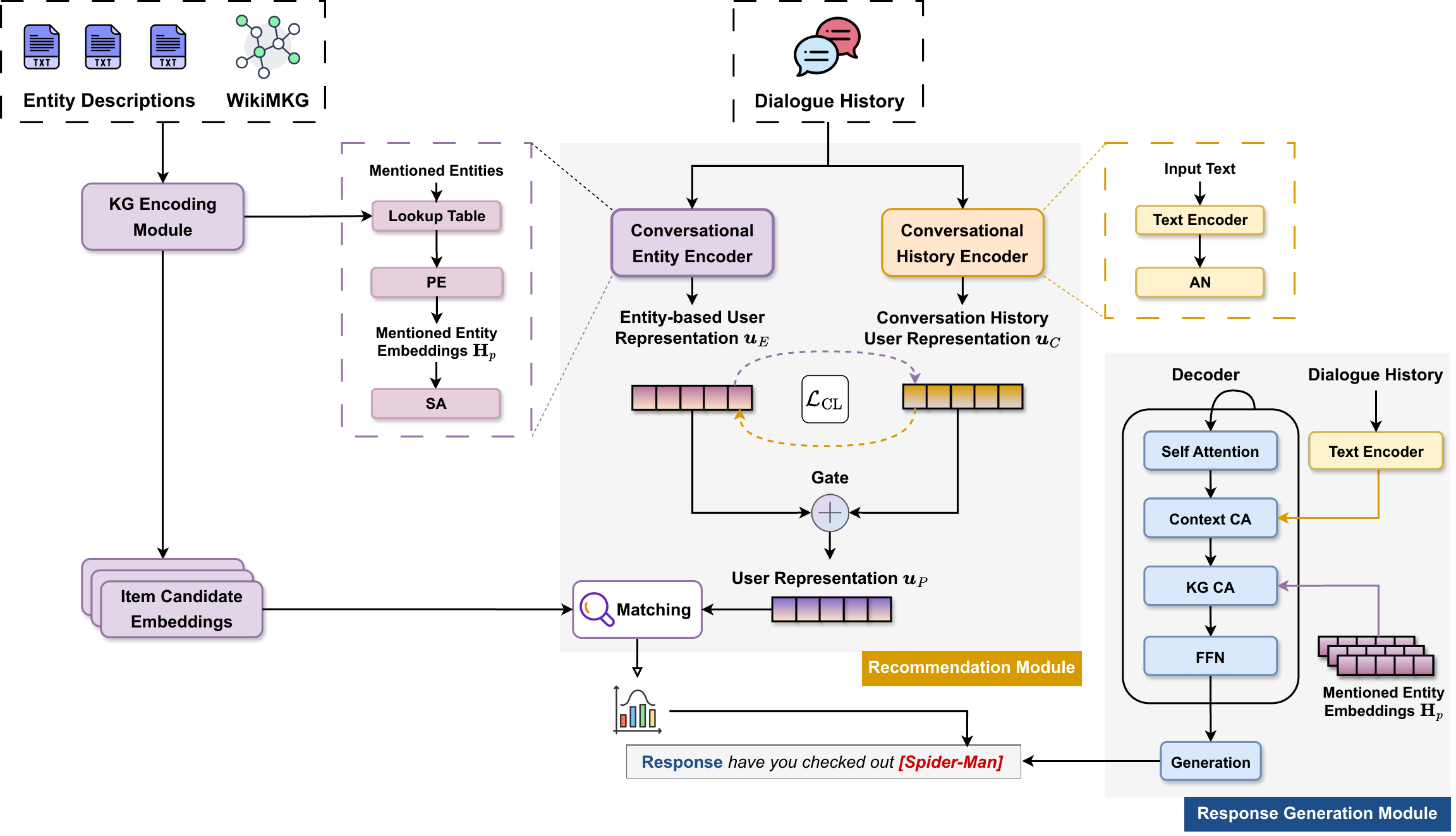}

\caption{The overview of the framework of the proposed KERL in a movie recommendation scenario. The Attention Network (AN) selectively focuses on relevant tokens. Positional Encoding (PE) and Self-Attention (SA) mechanisms preserve the sequence order and context, respectively. Context Cross-Attention (CA) and KG Cross-Attention integrate conversational and knowledge graph cues. The Recommendation Module matches items to user preferences, and the Response Generation Module formulates natural language suggestions.}

  \label{fig:model_overview}
\end{figure*}

\section{Methodology}\label{sec:methodology}
In this section, we present  \textbf{KERL} with its overview shown in Figure~\ref{fig:model_overview}, which consists of three main modules:
knowledge graph encoding module, recommendation module, and response generation module.
We first formalize the conversational recommendation task in Section~\ref{sec:formulation}, followed by a brief architecture overview in Section~\ref{sec:workflow}, 
we then introduce the process of encoding entity information (Section~\ref{sec:kg_encoder}), followed by our approach to both recommendation (Section~\ref{sec:rec}) and conversation tasks (Section~\ref{sec:gen}).
Finally, we present our training algorithm of KERL in Section~\ref{sec:training}.

\subsection{Problem Formulation}
\label{sec:formulation}

Formally, let $C = \{c_1, c_2, ..., c_m\}$ denote the history context of a conversation, where $c_m$ represents the utterance $c$ at the $m$-th round of a conversation.
Each utterance $c_m$ is either from the seeker (i.e., user) or from the recommender (i.e., system).
At the $m$-th round, the recommendation module will select items $\mathcal{I}_{m+1}$ from a set of candidate items $\mathcal{I}$ based on the estimated user preference, and the response generation module will generate a response $c_{m+1}$ to prior utterances.
Note that $\mathcal{I}_{t}$ can be empty when there is no need for a recommendation (i.e., chit-chat).

For the knowledge graph, let $\mathcal{G} = \{ (h, r, t) | h, t \in{\mathcal{E}}, r \in\mathcal{R} \}$ denote the knowledge graph, where each triplet $(h, r, t)$ describes a relationship $r$ between the head entity $h$ and the tail entity $t$. 
The entity set ${\mathcal{E}}$ contains all movie items in $\mathcal{I}$ and other non-item entities that are movie properties (i.e., director, production company).

\subsection{Architecture Overview}\label{sec:workflow}

This section elaborates on the workflow integrating the knowledge graph encoding module, the knowledge-enhanced recommendation module, and the knowledge-enhanced response generation module, as shown in Figure~\ref{fig:model_overview}. These components work together to process user inputs and generate personalized recommendations, as demonstrated in scenarios such as suggesting a superhero movie.

\begin{itemize}

    \item Knowledge Graph Encoding Module: This module integrates textual descriptions and entity relationships from the WikiMKG. When a user mentions an interest in \textit{superhero} films and watched \textit{Avengers: Infinity War}, this module engages by extracting and encoding relevant entity information (e.g. actors, directors, and movie characteristics). This process ensures a rich understanding of the entities, laying the foundation for contextually aware recommendations.

    \item Knowledge-enhanced Recommendation Module: This module synthesizes the encoded entity information with the user's conversational history, capturing the essence of the ongoing dialogue. For instance, when a user, after discussing \textit{Avengers: Infinity War}, seeks something different, this module can suggest a new superhero movie, aligning with their current preferences.

    \item Knowledge-enhanced Response Generation Module: This final module brings together the insights from the KG and the conversational context to formulate coherent and engaging responses. In our superhero movie scenario, it would generate a natural language suggestion that not only aligns with the user's expressed interest, but also fits seamlessly into the flow of the conversation. For example, the system suggests ``Have you checked out Spider-Man?'' as a fresh yet relevant choice.

\end{itemize}

\begin{figure}[t]
  \centering
  \includegraphics[width=\columnwidth]{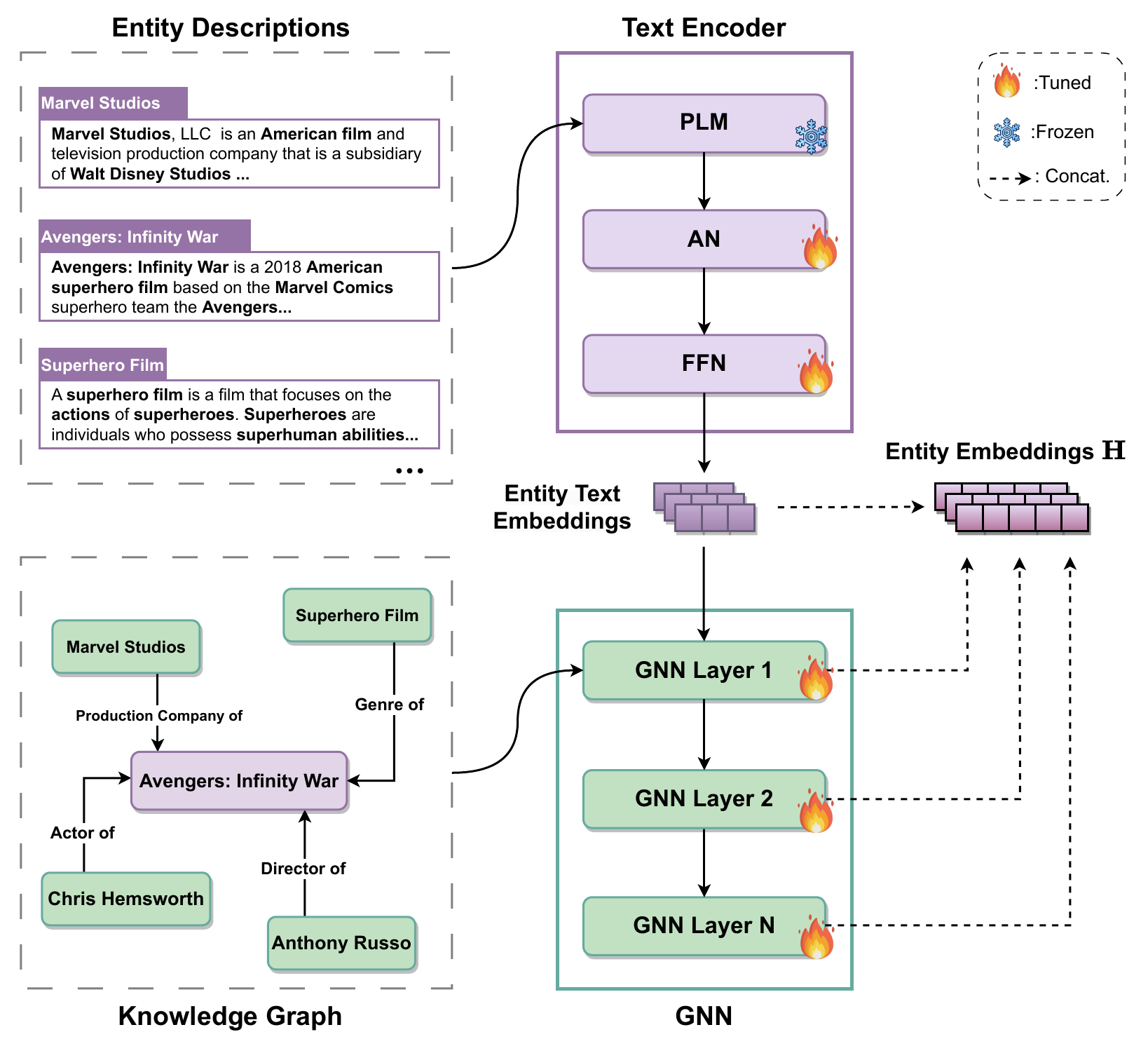}

\caption{The Knowledge Graph Encoding Module employs a PLM for textual semantics and a GNN for structural relations. It generates entity embeddings that include item embeddings, which are a subset used for recommendations. ``AN'' denotes the attention network.}
  \label{fig:kg_encoding}
\end{figure}
\subsection{Knowledge Graph Encoding Module}\label{sec:kg_encoder}
The Knowledge Graph Encoding Module is central to the proposed framework, as illustrated in detail in Figure~\ref{fig:kg_encoding}, designed to integrate rich textual descriptions with complex entity relationships from the KG. This integration is crucial, as it ensures that the system captures both the semantic and structural nuances of entities, providing a solid foundation for the CRS's contextual awareness and recommendation mechanisms.
\\

\subsubsection{PLM for Entity Encoding}

Previous studies focused on modeling entity representations based on their relationships with other entities, without taking into account the rich information contained in the text description of entities. To address this limitation, we propose to leverage the textual descriptions of entities to encode entities into vector representations.
Our innovative approach builds upon the capabilities of PLMs such as BERT~\cite{devlin_bert_2019}, BART~\cite{lewis-etal-2020-bart} and GPT2~\cite{Radford2019LanguageMA}. Within our framework, a domain-specific or generic PLM serves as a fundamental component for encoding textual information. These PLMs have demonstrated superior performance on various tasks in natural language processing. In our architecture, the entity text encoder is designed to be adaptable and customizable, ensuring its compatibility with a range of PLMs. For the purpose of this study, we use a pre-trained distillation BERT~\cite{iulia_well_2019} as our entity text encoder to capture the contextual word representations, underlining the adaptability and efficacy of our approach.

In the entity descriptions, different words carry varying degrees of significance in representing the entity. Therefore, computing the average of hidden word representations, which assigns equal weight to all words, may lead to a loss of crucial information. Thus, we adopt an attention network that integrates multi-head self-attention~\cite{vaswani_attention_2017}, and an attention-pooling block~\cite{wu-etal-2019-neural-news}. This approach allows the model to selectively focus on the important words in entity descriptions, facilitating the learning of more informative entity representations.
More specifically, for a given entity description text denoted as $[w_1, w_2, ..., w_k]$, with $k$ representing the total number of words in the description, we first transform the words into embeddings using the BERT model. These embeddings are fed into multiple Transformer~\cite{vaswani_attention_2017} layers to produce hidden word representations. We then use an attention network and a feed-forward network to summarize the hidden word representations into a unified entity embedding. The attention weights are calculated as follows:

\begin{equation}\label{eq:add-attention}
    \begin{aligned}
        \boldsymbol{\alpha}_i^{w} = \frac{\exp(\boldsymbol{q}_{w}^{\top} \cdot \sigma(\mathbf{V}_w \mathbf{w}_i + \mathbf{v}_w)}{\sum_{j=1}^{k} \exp(\boldsymbol{q}^{\top}_{w} \cdot \sigma(\mathbf{V}_w \mathbf{w}_j + \mathbf{v}_w)}
    \end{aligned}
\end{equation}
where $\boldsymbol{\alpha}_i^{w}$ denotes the attention weights of the $i$-th word in the entity descriptions, $\mathbf{w}_i$ denotes the word representations obtained from the top layer of BERT, $\sigma$ is tanh activation function, $\mathbf{V}_w$ and $\mathbf{v}_w$ are  projection parameters, and $\boldsymbol{q}_{w}$ is the query vector. To generate the entity embeddings, we aggregate the representation of these word representations by employing the weighted summation of the contextual word representation and a feed-forward network, formulated as:

\begin{equation}\label{eq:add-attention-ffn}
    \mathbf{h}_e^{d} = \textrm{FFN}\left(\sum_{i=1}^{k} \boldsymbol{\alpha}_i \mathbf{x}_i\right)
\end{equation}
where $\textrm{FFN}(\cdot)$ denotes a fully connected feed-forward network, consisting of two linear transformations with an activation layer, $\mathbf{h}_e^{d}$ represents the entity embedding that captures the rich information contained in its textual description. We omit equations related to BERT and multi-head attention. \newline

\subsubsection{Knowledge Graph Embedding}

In addition to the textual descriptions of entities, semantic relationships between entities can provide valuable information and context.
To capture this information, we adopt Relational Graph Convolutional Networks (R-GCNs)~\cite{schichtkrull_modeling_2018,chen-etal-2019-towards,zhou_c-crs_2022} to encode structural and relational information between entities. Formally, the representation of the entity $e$ at the $(\ell+1)$-th layer is calculated as follows:

\begin{equation}\label{eq:rgcn}
    \begin{aligned}
        \mathbf{h}_{e}^{(\ell+1)} = \sigma \left( \sum_{r\in\mathcal{R}}\sum_{e^{\prime}\in{\mathcal{E}_{e}^{r}}}\frac{1}{Z_{e,r}}\mathbf{W}_{r}^{(\ell)}\mathbf{h}_{e^{\prime}}^{(\ell)} + \mathbf{W}_{e}^{(\ell)}\mathbf{h}_{e}^{(\ell)} \right)
    \end{aligned}
\end{equation}
where $\mathbf{h}_{e}^{\ell}$ is the embedding of entity $e$ at the $\ell$-th layer and $\boldsymbol{h}_{e}^{0}$ is the textual embedding of the entity derived from the text encoder described in Equation~\ref{eq:add-attention-ffn}. The set ${\mathcal{E}_{e}^{r}}$ consists of neighboring entities that are connected to the entity $e$ through the relation $r$. $\mathbf{W}_{r}^{(\ell)}$ and $\mathbf{W}_{e}^{(\ell)}$ are learnable model parameters matrices,~${Z_{e,r}}$ serves as a normalization factor, and $\sigma$ is the activation function.
As the output of different layers represents information from different hops, we use the layer-aggregation mechanism ~\cite{xu2018representation} to concatenate the representations from each layer into a single vector as:

\begin{equation}\label{eq:kg_agg}
    \begin{aligned}
        \mathbf{h}_{e}^{*} = \mathbf{h}_{e}^{(0)} || \cdots || \mathbf{h}_{e}^{(\ell)}
    \end{aligned}
\end{equation}
where $||$ is the concatenation operation. By doing so, we can enrich the entity embedding by performing the propagation operations as well as control the propagation strength by adjusting $\ell$. Finally, we can obtain the hidden representation of all entities in $\mathcal{G}$, which can be further used for user modeling and candidate matching.

To preserve the relational reasoning within the structure of the knowledge graph, we employ the TransE \cite{bordes_2013_translating} scoring function as our knowledge embedding objective to train our knowledge graph embedding. This widely used method learns to embed each entity and relation by optimizing the translation principle of $h + r \approx t$ for a valid triple ($h$, $r$, $t$).
Formally, the score function $d_r$ for a triplet $(h, r, t)$ in the knowledge graph $\mathcal{G}$ is defined as follows:

\begin{equation}\label{eq:transE}
    \begin{aligned}
        d_r(h, t) & = ||\mathbf{e}_h + \mathbf{r} - \mathbf{e}_t ||_{p} \\
    \end{aligned}
\end{equation}
where $h$ and $t$ represent the head and tail entities, respectively.
$\mathbf{e}_h$ and $\mathbf{e}_t$ are the new entity embeddings from Equation~\ref{eq:kg_agg}, and $\mathbf{r} \in \mathbb{R}^{|\mathcal{R}|\times d}$ are the embeddings of the relation, $d$ is the embedding dimension, and $p$ is the normalization factor.
The training process of knowledge graph embeddings prioritizes the relative order between valid and broken triplets and encourages their discrimination through margin-based ranking loss~\cite{sun2018rotate}:

\begin{equation}\label{eq:transEloss}
    \begin{aligned}
        \mathcal{L}_{\text{KE}}  = & - \log\sigma(\gamma - d_r(h, t))                                          \\
                                   & - \sum_{i=1}^{k}\frac{1}{k}\log\sigma(d_r(h_{i}^{'}, t_{i}^{'}) - \gamma)
    \end{aligned}
\end{equation}
where $\gamma$ is a fixed margin, $\sigma$ is the sigmoid function, and $(h_{i}^{'}, r, t_{i}^{'})$ are broken triplets that are constructed by randomly corrupting either the head or tail entity of the positive triples $k$ times.

\subsection{Knowledge-enhanced Recommendation Module}\label{sec:rec}

The Knowledge-enhanced Recommendation Module synthesizes entity-based user representation with contextual dialogue history to deliver precise and context-aware recommendations. This module employs a conversational entity encoder, integrating positional encoding with self-attention to reflect the sequential significance of entities in dialogue. Complementing this, a conversational history encoder captures the textual nuances of user interactions. A contrastive learning strategy harmonizes these dual representations, sharpening the system's ability to discern and align with user preferences for more personalized recommendations.
\newline

\subsubsection{Conversational Entity Encoder}

To effectively capture user preferences with respect to the mentioned entities, we employ positional encoding and a self-attention mechanism \cite{LinFSYXZB17}.
First, we extract non-item entities and item entities in the conversation history $C$ that are matched to the entity set $\mathcal{E}$. Here, item entities refer to movie items, while non-item entities pertain to movie properties such as the director and production company.
Then, we can represent a user as a set of entities $\mathbf{E}_u = \{e_1, e_2, ..., e_i\}$, where $e_i \in {\mathcal{E}}$. After looking up the entities in $\mathbf{E}_{u}$ from the entity representation matrix $\mathbf{H}$ (as shown in Equation~\ref{eq:kg_agg}), we obtain the respective embeddings $\mathbf{H}_{u}=(\mathbf{h}_{1}^{*}, \mathbf{h}_{2}^{*}, ..., \mathbf{h}_{i}^{*})$.

Previous research~\cite{chen-etal-2019-towards, zhou_improving_2020, lu_revcore_2021} has primarily relied on the self-attention mechanism~\cite{LinFSYXZB17} to summarize the user's preference over the mentioned entities.
However, such an approach ignores the order of entities within a conversation, which can significantly affect the topic of a conversation, leading to different conversational trajectories.
Therefore, we use the learnable positional encoding inspired by the Transformer~\cite{vaswani_attention_2017} architecture.
This approach is commonly used in NLP to encode information about the position of a token in a sentence, allowing the model to understand the relationship between tokens based on their position in a sentence.
We first utilize it to capture the order of entity appearance within the conversation, ensuring more accurate summaries of user preferences.
Formally, the entity-level representation of the user that includes positional information can be formulated as follows:
\begin{equation}\label{eq:pe}
    \begin{aligned}
        \mathbf{H}_{p} & =   \mathbf{H}_{u} + \textbf{E}_{pos},
    \end{aligned}
\end{equation}
where $\mathbf{H}_{p}$ denotes the entity-level representation of the user that considers entity positional information.
The positional encoding matrix, denoted as $\textbf{E}_{pos}$, has dimensions $\in\mathbb{R}^{|P|\times{d}}$. Here, $|P|$ corresponds to the length of the entity sequence, and $d$ denotes the embedding dimension.
The positional encoding matrix is added to the user entity representation set $\mathbf{H}_{u}$ based on the corresponding order of appearance of the entity.
Then, we adopt the self-attention mechanism~\cite{LinFSYXZB17} to summarize the entity-based user representation $\boldsymbol{u}_{E}$ as follows:
\begin{equation}\label{eq:self-att}
    \begin{aligned}
        \boldsymbol{u}_{E}      & = \mathbf{H}_{p} \cdot \boldsymbol{\alpha^{e}}                                          \\
        \boldsymbol{\alpha^{e}} & = \textrm{softmax}(\boldsymbol{b}_{e}^{\top} \cdot \sigma (\mathbf{W}_p\mathbf{H}_{p})) \\
    \end{aligned}
\end{equation}
where $\boldsymbol{\alpha^{e}}$ denotes the attention weight vector that reflects the importance of each interacted entity, and $\mathbf{W}_p$ and $\boldsymbol{b}_{e}$ are learnable parameter matrix and vector, and $\sigma$ is the tanh activation function. 
Finally, we can derive the entity-based user representation~$\boldsymbol{u}_{E}$ that considers the importance of each interacted entity and their chronological context within the conversation history, providing a nuanced and temporally-informed representation of user preferences at the entity level.
\newline

\subsubsection{Conversational History Encoder}

Despite the capability of knowledge-enhanced entity-based user representation to model user preferences based on entities mentioned in context $C$, it can lead to misinterpretations of user preferences, as it did not take into account the actual textual content of the conversation. For example, consider a scenario where a user states ``I am not a fan of the movie A''. This negative sentiment towards ``movie A'' cannot be accurately captured using only entity-level representation. Therefore, we employ a conversational history encoder to obtain a representation that encompasses the textual content of the conversation.
Specifically, a conversation history $C$ is composed of $m$ utterances that are generated by the user and the recommender. We concatenate the dialogue utterance $c_m$ in chronological order to create an extended sentence $\{c_1; c_m \}$. To compute the contextual word representation, we employ the BART \cite{lewis-etal-2020-bart} encoder as our conversational history encoder and denote it as:

\begin{equation}\label{eq:bart-encoder}
    \begin{aligned}
        \mathcal{F^C} = \text{BART-Enc}(c_1; c_m)
    \end{aligned}
\end{equation}
where $\mathcal{F^C}$ represents contextual word representations. Next, we utilize the same attention network as in Equation~\ref{eq:add-attention} and Equation~\ref{eq:add-attention-ffn} to derive the conversation history user representation~$\boldsymbol{u}_C$. This approach enables us to capture the contextual-level user preferences, which can complement entity-level preferences.
\newline

\subsubsection{Semantic Alignment via Contrastive Learning}

To effectively leverage both conversation history user representation and entity-based user representation, it is important to fuse them in a way that captures the user's preferences from both perspectives.

Previous works~\cite{zhou_improving_2020, zhou_c-crs_2022} have shown that aligning the different types of representations can significantly improve the performance of the CRS. Therefore, we adopt the contrastive learning framework~\cite{chen_2020_simple, zhou_c-crs_2022} to bridge the gap between the conversation history user representation~$\boldsymbol{u}_{C}$ and the entity-based user representation~$\boldsymbol{u}_{E}$.
More specifically, we take pairs $(\boldsymbol{u}_{C},\boldsymbol{u}_{E})$ as positive samples, representing the same user's preferences from two distinct views: the conversational context and their entity interactions. In contrast, the representations from different users in the same batch are treated as negative examples. We compute the contrastive loss as:

\begin{equation}\label{eq:cl}
    \begin{aligned}
        \mathcal{L}_{\text{CL}} & = \log\frac{\text{exp}(\text{sim}{(\mathbf{z}, \textbf{z}^+)/ \tau})}{\sum_{\textbf{z}_{i}^{-}\in\{{\textbf{z}^{-}}\}}\text{exp}(\text{sim}{(\mathbf{z}, \textbf{z}_{i}^-)/ \tau})}
    \end{aligned}
\end{equation}
where $\mathbf{z}$ and $\mathbf{z}^+$ are positive pairs, $\{\mathbf{z}^-\}$ is the negative example set for $\mathbf{z}$, $\text{sim}(\cdot)$ is the cosine similarity function, and $\tau$ is a temperature hyperparameter. By leveraging contrastive learning objectives, we can pull together representations from different views, and enable them to mutually improve and contribute to a more comprehensive representation of the user's preferences.
\newline

\subsubsection{Entity-Context Fusion for Recommendation}

To incorporate both entity-based user preferences and conversation history user representation, we utilize a gate mechanism to derive the preference representation $\boldsymbol{u}_{P}$ of the user $u$:
\begin{equation}\label{eq:gate}
    \begin{aligned}
        \boldsymbol{u}_{P} & = \beta \cdot \boldsymbol{u}_{E} + (1 - \beta) \cdot \boldsymbol{u}_{C}                                & \\
        \beta              & =  \textrm{sigmoid}\Big(\mathbf{W}_{\textrm{gate}}(\boldsymbol{u}_{E}  \;\| \; \boldsymbol{u}_{C}\Big) &
    \end{aligned}
\end{equation}
where ${\beta}$ represents the gating probability, $\mathbf{W}_{\textrm{gate}}$ is learnable parameters matrix. Finally, the probability of recommending an item $i$ from the item set $\mathcal{I}$ to the user $u$ is calculated as follows:

\begin{equation}\label{eq:rec}
    \begin{aligned}
        \text{P}_{\text{rec}}(i) = \textrm{softmax}(\boldsymbol{u}_{P}^{\top} \cdot \mathbf{H}_i),
    \end{aligned}
\end{equation}
where $\mathbf{H}_i$ is the learned item embedding for item $i$. To learn the parameters, we use cross-entropy loss as the optimization objective:

\begin{equation}\label{eq:recloss}
    \begin{aligned}
        \mathcal{L}_{\text{rec}} & = - \sum^{M}_{j=1}\sum^{N}_{i=1} y_{ij} \cdot \text{log}\Big(\text{P}_{\text{rec}}^{(j)}(i)\Big)
    \end{aligned}
\end{equation}
where $M$ is the number of conversations, $j$ is the index of the current conversation, $N$ is the number of items, and $i$ is the index of items. $y_{ij}$ denotes the ground-truth label.

\subsection{Knowledge-enhanced Response Generation Module}\label{sec:gen}

In this section, we study how to fine-tune the proposed response generation module to produce informative responses, utilizing the BART \cite{lewis-etal-2020-bart} model. The BART model is a pre-trained language model that is built upon on Transformer \cite{vaswani_attention_2017} architecture and trained using denosing objectives. It has been demonstrated to be effective in many natural language generation tasks, including summarisation, machine translation, and abstractive QA. Our implementation of the encoder adheres to the standard BART Transformer architecture, and we devote our attention to introducing the knowledge-enhanced decoder.

To enhance response generation during decoding, we adopt a approach similar to KGSF \cite{zhou_improving_2020}. Specifically, we integrate knowledge-enhanced representation of entities, as described in Equation~\ref{eq:pe},
and fused via attention layers as follows:

\begin{equation}\label{eq:self-attention}
    \begin{aligned}
        \mathbf{A}_{0}^{l} & = \textrm{MHA}(\mathbf{Y}^{l-1}, \mathbf{Y}^{l-1}, \mathbf{Y}^{l-1}) \\
        \mathbf{A}_{1}^{l} & = \textrm{MHA}(\mathbf{A}_{0}^{l}, \mathbf{X}, \mathbf{X})           \\
        \mathbf{A}_{2}^{l} & = \textrm{MHA}(\mathbf{A}_{1}^{l}, \mathbf{H}_p, \mathbf{H}_p)       \\
        \mathbf{Y}^{l}     & = \text{FFN}(\mathbf{A}_{2}^{l}),                                    \\
    \end{aligned}
\end{equation}
where $\textrm{MHA}(\mathbf{Q},\mathbf{K},\mathbf{V})$ denotes the multi-head attention function \cite{vaswani_attention_2017}, which takes a query matrix ~$\mathbf{Q}$, key matrix ~$\mathbf{K}$ and value matrix ~$\mathbf{V}$ as input. $\textrm{FFN}(\cdot)$ represents a fully connected feed-forward network.
$\mathbf{X}$ represents the context token embeddings from the BART encoder, and $\mathbf{H}_p$ corresponds to the knowledge-enhanced entity embeddings obtained from the recommendation module. $\mathbf{A}_{0}^{l}$, $\mathbf{A}_{1}^{l}$, $\mathbf{A}_{2}^{l}$ refer to the representations after self-attention, cross-attention with encoder output, cross-attention with knowledge-enhanced entity embedding, respectively.
By following this chain, our decoder is capable of injecting valuable knowledge information from the entity description and knowledge graph, resulting in more informative and relevant responses.
Finally, the decoder generates a matrix of representations, denoted by $\mathbf{Y}^{l}$ at the $l$-th layer.

Moreover, we employ the copy mechanism \cite{gu-etal-2016-incorporating} to enhance the generation of tokens associated with entities:
\begin{equation}\label{eq:copy}
    \begin{aligned}
        \text{Pr}({y_i} | \{y_{i-1}\}) = \text{Pr}_{1}(y_i|\mathbf{Y}_{i}) + \text{Pr}_{2}(y_i|\mathbf{Y}_{i}, \mathcal{G})
    \end{aligned}
\end{equation}
where $\text{Pr}({y_i} | \{y_{i-1}\})$ is the generative probability over the vocabulary based on the last generated sequence $\{y_{i-1}\}=[y_1,\cdots,y_i]$.
$\text{Pr}_{1}(\cdot)$ is a generative probability with the decoder output $\mathbf{Y}_{i}$, and $\text{Pr}_{2}(\cdot)$ is the copy probability of the tokens from the entity descriptions in the knowledge graph $\mathcal{G}$. Both probability functions are implemented with a softmax operation. To learn the response generation module, we use a cross-entropy loss:

\begin{equation}\label{eq:gen-loss}
    \begin{aligned}
        \mathcal{L}_{\text{gen}} = \frac{1}{M}\sum_{m=1}^{M}\text{log}(\text{Pr}(c_m|c_1,..., c_{m-1})))
    \end{aligned}
\end{equation}
where $M$ is number of turn in a conversation $C$, $c_m$ denotes the $t$-th turns in the conversation.

\begin{algorithm}[t]
    \small
    \caption{The training algorithm of KERL.}\label{sec:parameter training}
    \LinesNumbered
    \KwIn{The conversational recommendation dataset $\mathcal{X}$ and knowledge graph $\mathcal{G}$}
    \KwOut{Model parameters $\Theta_{G}$, $\Theta_{R}$ and $\Theta_{C}$.}
    Pre-train the $\Theta_{G}$ and $\Theta_{R}$ by minimizing losses $\mathcal{L}_{\text{KE}}$ and $\mathcal{L}_{\text{CL}}$ in Eq.~\ref{eq:transEloss} and Eq.~\ref{eq:cl}, respectively. \\
    \For{$t = 1 \to |\mathcal{X}|$}{
    Acquire textual embeddings $\mathbf{h}_e^{d}$ from entity descriptions in $\mathcal{G}$ by Eq.~\ref{eq:add-attention-ffn}. \\
    Acquire entity embeddings from $\mathcal{G}$ by Eq.~\ref{eq:rgcn}. \\
    Determine entity-based user preferences $\boldsymbol{u}_E$ by Eq.~\ref{eq:pe} and Eq.~\ref{eq:self-att}. \\
    Acquire conversation history user representation $\boldsymbol{u}_C$ by Eq.~\ref{eq:bart-encoder} and Eq.~\ref{eq:add-attention-ffn}. \\
    Compute final user representation $\boldsymbol{u}_{P}$ by gate mechanism using Eq.~\ref{eq:gate}. \\
    Compute $\text{P}_{\text{rec}}(i)$ using Eq.~\ref{eq:rec}. \\
    Update $\Theta_{G}$ and $\Theta_{R}$ as per Eq.~\ref{eq:recloss}.
    }
    \For{$i = 1 \to |\mathcal{X}|$}{
    Acquire textual embeddings $\mathbf{h}_e^{d}$ from entity descriptions in $\mathcal{G}$ by Eq.~\ref{eq:add-attention-ffn}. \\
    Acquire entity embeddings from $\mathcal{G}$ by Eq.~\ref{eq:rgcn} and use Eq.~\ref{eq:pe} to add entity positional information.  \\
    Acquire conversation history word representation $\mathcal{F^C}$ by Eq.~\ref{eq:bart-encoder}. \\
    Acquire $\mathbf{Y}^{l}$ by BART decoder using Eq.~\ref{eq:self-attention}. \\
    Compute $\text{Pr}({y_i} | \{y_{i-1}\})$ based on Eq.~\ref{eq:copy}. \\
    Update $\Theta_{C}$ as per Eq.~\ref{eq:gen-loss}.
    }
    \Return $\Theta_{G}$, $\Theta_{R}$ and $\Theta_{C}$
\end{algorithm}
\subsection{Parameter Learning}\label{sec:training}

The KERL consists of three modules, namely, the knowledge graph encoding module, the recommendation module, and the response generation module, which are denoted by the parameters $\Theta_{G}$, $\Theta_{R}$ and $\Theta_{C}$, respectively. These modules collectively determine the training algorithm for KERL, as detailed in Algorithm~\ref{sec:parameter training}.

The initial stage of the training involves the pre-training of the knowledge graph encoding module $\Theta_{G}$ and the recommendation module $\Theta_{R}$ using margin-based ranking loss $\mathcal{L}_{\text{KE}}$ and contrastive loss $\mathcal{L}_{\text{CL}}$.
Subsequently, the parameters of these modules are explicitly optimized for the recommendation task.
Every iteration begins by exploiting WikiMKG to obtain the corresponding entity embeddings using Equation~\ref{eq:add-attention-ffn} and Equation~\ref{eq:rgcn}.
Positional encoding and self-attention are then carried out to determine the entity-based user preferences.
Meanwhile, the conversation history user representation is obtained using Equation~\ref{eq:bart-encoder} and Equation~\ref{eq:add-attention-ffn}.
Then, we perform the gate mechanism to infer the final user representation.
Finally, we compute the cross-entropy loss using Equation~\ref{eq:recloss}, followed by applying gradient descent to update parameters $\Theta_{G}$ and $\Theta_{R}$.

We start optimizing the parameters in $\Theta_{C}$ once the loss of the recommendation component reaches convergence. At each iteration, we still have to obtain the entity embedding through WikiMKG first. Then, the relevant dialogue context is encoded using the BART encoder and decoder. After calculating the generation probability using Equation~\ref{eq:copy}, we compute the cross-entropy loss using Equation~\ref{eq:gen-loss}, followed by applying gradient descent to update parameters $\Theta_{C}$.

\section{Experiment Setup}\label{sec:experiment}

\subsection{Dataset}
\label{sec:dataset}

\begin{table}[]
\centering
\caption{Statistics of the datasets after preprocessing.}
\label{tab:dataset}
\begin{tabular}{crrr}
\toprule
Dataset  & \multicolumn{1}{c}{\textbf{\#Dialogues}} & \multicolumn{1}{c}{\textbf{\#Utterances}} & \multicolumn{1}{c}{\textbf{\#Unique Items}} \\ \midrule
ReDial   & 11,348                                    & 205,644                                   & 6,891                                      \\
Inspired & 1,001                                     & 35,686                                     & 1,866                            \\ \bottomrule              
\end{tabular}%

\end{table}

Experiments were conducted using two widely recognized datasets: ReDial ~\cite{li_towards-deep_2018} and INSPIRED ~\cite{hayati-etal-2020-inspired}.
ReDial is the most commonly used dataset for conversational recommendation. This dataset was built using Amazon Mechanical Turk (AMT), where pairs of workers act as movie seekers and recommenders, following comprehensive instructions. 
INSPIRED is another CRS dataset for movies, also constructed using AMT. It features annotated recommendation strategies based on social science theories. The statistics of both datasets are summarized in Table \ref{tab:dataset}.
\subsection{Knowledge Graph construction}
\label{sec:build-kg}

Existing research~\cite{chen-etal-2019-towards, zhou_improving_2020, sarkar-etal-2020-suggest, zhou_c-crs_2022} has mainly focused on using open-domain KGs such as DBpedia and ConceptNet. 
However, these KGs may include many irrelevant entities~\cite{zhang-etal-2022-toward} and do not incorporate entity descriptions. 
To address these limitations, we propose the Wiki Movie Knowledge Graph (WikiMKG) that incorporates entity descriptions from Wikipedia. 
Specifically, we collect movie information from WikiData\footnote{\url{https://www.wikidata.org}}, a multilingual knowledge graph that is collaboratively edited, along with corresponding descriptions from Wikipedia. 
We use movie names and release years as keywords to search on Wikidata and keep movie properties such as genres, cast members, directors, and production companies. 
\subsection{Baselines}
\label{sec:baseline}

To demonstrate effectiveness, we evaluate our proposed KERL on recommendation and conversation tasks. To ensure a comprehensive evaluation, we do not limit our comparison to existing CRS methods, but also representative approaches from both the recommendation and conversation domains as well.

\begin{itemize}
  \item \textbf{Popularity}: This method ranks items according to their frequency of recommendation in the training set.
  \item \textbf{TextCNN}~\cite{kim-2014-convolutional}: It uses CNN model to extract user preferences from conversational context to rank items.
  \item \textbf{Transformer}~\cite{vaswani_attention_2017}: It applies a Transformer-based encoder-decoder method to generate responses.
  \item \textbf{BERT}~\cite{devlin_bert_2019}: It is a pre-trained model based on the Transformer architecture, using the masked language modeling task on a large-scale corpus. We use the representation of the [CLS] token for recommendation, which is a common practice in language models.
  \item \textbf{GPT-2}~\cite{Radford2019LanguageMA}: It is an auto-regressive model that uses the Transfomer decoder. We use historical utterances as the input to generate the response.
  \item \textbf{ReDail}~\cite{li_towards-deep_2018}: This model consists of a recommender module based on auto-encoder architecture~\cite{he_distributed-representation_2017}, and a response generation module based on the hierarchical recurrent encoder-decoder~\cite{serban_hierarchical_2017}.
  \item \textbf{KBRD}~\cite{chen-etal-2019-towards}: This model utilizes DBpedia to enhance the semantics of contextual items or entities for the recommendation task. The response generation module is based on the Transformer architecture \cite{vaswani_attention_2017} and adopts KG information to assist generation.
  \item \textbf{KGSF}~\cite{zhou_improving_2020}: This method uses mutual information maximization to align the semantic space of entity-oriented and word-oriented KGs for the recommendation task. The response generation module follows a Transformer encoder~\cite{vaswani_attention_2017} and a KG fused Transformer decoder.
  \item \textbf{C$^2$-CRS}~\cite{zhou_c-crs_2022}: This method uses contrastive learning to align different types of data to generate a more coherent fusion representation.
  \item \textbf{UniCRS}~\cite{wang_towards_2022}: This method utilizes a pre-trained language model and unifies the recommendation and conversation task into the prompt learning paradigm.
\end{itemize}

In these baselines, Popularity, TextCNN~\cite{kim-2014-convolutional} and BERT~\cite{devlin_bert_2019} are recommendation methods; Transformer~\cite{vaswani_attention_2017} and GPT-2~\cite{Radford2019LanguageMA} are text generation methods. As there are no records of user-item interaction, except dialogue utterances, we do not include other recommendation models. Furthermore, ReDial~\cite{li_towards-deep_2018}, KBRD~\cite{chen-etal-2019-towards}, KGSF~\cite{zhou_improving_2020}, C$^2$-CRS~\cite{zhou_c-crs_2022} and UniCRS~\cite{wang_towards_2022} are conversational recommendation methods.

\subsection{Evaluation Metrics}
\label{sec:metrics}

We use different metrics to evaluate the recommendation and conversation tasks. For the recommendation task, we adopt Recall@$K$ (R@$K$, $K$ = 1, 10, 50) for evaluation. Recall@$K$ indicates whether the ground truth label is in the list of predicted top-k items. Response generation evaluation consists of automatic and human evaluations. Following previous CRSs work \cite{chen-etal-2019-towards, zhou_improving_2020, wang_towards_2022}, we adopt Distinct-n approach (Dist-n, n = 2, 3, 4) as the automatic evaluation to assess the diversity of the generated responses at the sentence level. This metric is defined as the ratio of unique n-gram to the total number of sentences. To evaluate the informativeness of responses, we calculate the item ratio \cite{zhou_improving_2020, zhou_crfr_2021, wang_finetuning_2021}, which is the proportion of responses containing items. In contrast to traditional conversation tasks, our approach does not aim to generate responses that mirror the ground truth. As such, we do not employ metrics like BLEU, which measure the similarity between the ground truth and the generated utterances. This approach aligns with the guidelines set forth by \cite{zhou_improving_2020, lu_revcore_2021, zhou_c-crs_2022}. Instead, we adopt a human evaluation approach. Annotators assess the generated responses, which are randomly selected from 100 multi-turn dialogues from the test set. The evaluation is based on two criteria: fluency and informativeness. Fluency assesses the smoothness and coherence of generated responses, determining whether they follow a logical and grammatically correct structure. Informativeness, on the other hand, examines whether the responses contain rich information about the recommended items. The scoring scale for both criteria ranges from 0 to 2, with 0 indicating poor performance and 2 indicating excellent performance. The average results for all the above metrics are reported.
\subsection{Implementation Details}
\label{sec:implementation}

We implemented KERL in PyTorch \cite{Paszke2017AutomaticDI}, training it efficiently on a single NVIDIA Tesla A100 card.
We utilized a fixed BERT-mini~\cite{iulia_well_2019} model as the entity text encoder, which has 4 layers, 256 hidden units, and 4 attention heads, with 11.3M parameters. 
We froze all parameters in the backbone throughout the entire training process and only fine-tuned the attention network. Additionally, to further enhance training efficiency, we used a caching strategy wherein we precompute and store the hidden states from BERT-mini's lower layers for all entities in the WikiMKG prior to the training.

The maximum input length for the entity text encoder was set to 40, and the number of layers to 2 for R-GCN. The temperature $\tau$ in Equation~\ref{eq:cl} is set to 0.07.
We adopted the BART-Base~\cite{lewis-etal-2020-bart} model in the present study, which consists of six layers of encoder and decoder, respectively. 
We strategically deployed two independent BART encoders within distinct modules. The first BART encoder, fine-tuned for tasks related to recommendation, serves as our conversational history encoder, with the fine-tuning applied specifically to the last two layers.
Another BART encoder operates within the response generation module, with its final layer fine-tuned to excel in the generation task. All layers of the BART decoder were fine-tuned for the generation task.
During the inference phase, to maintain computational efficiency, we precompute and cache knowledge-enhanced entity embeddings. These precomputed embeddings are then efficiently retrieved as needed, significantly reducing the computational load during inference.
The generation module in this phase generates either the special token, which acts as a placeholder designated to be populated by items suggested by the recommendation module, or a general token from the original vocabulary. 
All placeholders will be filled with the recommendation items, effectively completing the response generation process.

In terms of training specifics, the knowledge embedding objective (Equation \ref{eq:transEloss}) used a batch size of 512 and a negative sample size of 128 for both datasets. For the recommendation task, we maintained the batch size of 128, with learning rates for BART and other components at 2e-5 and 6e-4, respectively. For the conversation task,  the batch size was 128, and learning rates were 1e-4 for ReDial and 2e-4 for INSPIRED. Early stopping was performed for the recommendation and conversation tasks based on the development set evaluation metric. All baseline models were implemented using the open-source toolkit CRSLab~\cite{zhou-etal-2021-crslab}, which provides various conversational recommendation models and benchmark datasets for comprehensive evaluations.

\section{Experimental Results}\label{sec:results}

The following section presents details of the experimental results.

\begin{table}[]
    \centering
    \caption{Evaluation results on the recommendation task. The best results are in bold, the second-best results are underlined. ``Improv." denotes the relative improvement compared to the second-best results.}
    \label{tab:recbase}
    \renewcommand{\arraystretch}{1.3}
    \resizebox{\linewidth}{!}{%
    \begin{tabular}{lcccccc}
        \toprule
            Dataset                & \multicolumn{3}{c}{ReDial} & \multicolumn{3}{c}{INSPIRED} \\ \midrule
            \multirow{2}{*}{Model} & \multicolumn{6}{c}{Recall@K}                              \\ \cmidrule(l){2-7} 
                                   & K=1    & K=10    & K=50    & K=1     & K=10     & K=50    \\ \midrule
            Popularity             & 0.012 & 0.061 & 0.179 & 0.032 & 0.155 & 0.323 \\
            TextCNN                & 0.013 & 0.068 & 0.191 & 0.047 & 0.161 & 0.327 \\
            BERT                   & 0.028 & 0.143 & 0.319 & 0.044 & 0.179 & 0.328 \\ \midrule
            ReDial                 & 0.024 & 0.140 & 0.320 & 0.030 & 0.117 & 0.285 \\
            KBRD                   & 0.031 & 0.150 & 0.336 & 0.058 & 0.172 & 0.265 \\
            KGSF                   & 0.038 & 0.183 & 0.378 & 0.064 & 0.175 & 0.273 \\
            C$^2$-CRS              & \underline{0.050} & \textbf{0.218} & 0.402 & 0.089 & 0.240 & 0.382 \\ 
            UniCRS                 & 0.048 & 0.216 & \underline{0.416} & \underline{0.091} & \underline{0.250} & \underline{0.408} \\ \midrule
            \textbf{KERL}          & $\textbf{0.056}$  & $\underline{0.217}$ & $\textbf{0.426}$ & $\textbf{0.106}$  & $\textbf{0.281}$ & $\textbf{0.439}$ \\ 
            Improv.              & 12.0\% & -0.1\% & 2.4\% & 16.5\% & 12.4\% & 7.6\% \\
        \bottomrule
    \end{tabular}%
}
\end{table}

\subsection{Evaluation on Recommendation Task}

Table \ref{tab:recbase} summarizes the overall performance of the various recommendation models. For baseline models, context-aware recommendation models (TextCNN and BERT) outperform Popularity, indicating that contextual information can be more beneficial for recommendation than general item popularity. Also, BERT outperformed TextCNN. This is likely due to BERT's pre-training and bidirectional transformer architecture, which allows it to capture the semantic relationships between words more effectively than TextCNN.

Among conversational recommendation methods, KBRD outperforms ReDial, as it uses the knowledge graph as external information to improve user preference modeling.
By introducing both entity-oriented knowledge graphs and word-oriented knowledge graphs, KGSF outperforms KBRD and ReDial.
C$^2$-CRS and UniCRS introduce advanced model architectures to further enhance performance.~C$^2$-CRS employs contrastive learning techniques to synergize information from multiple sources, enriching the representation of user preferences. UniCRS leverages a prompt-based learning framework, simplifying the integration of user preferences into the recommendation process.

The proposed KERL model shows superior performance on the ReDial and INSPIRED datasets, outperforming baselines and state-of-the-art systems with notable margins. This success is primarily due to its innovative combination of topological and textual information for enhanced entity representation. The inclusion of positional encoding and a conversational history encoder further refines user preference insights. In addition, the application of contrastive learning effectively aligns various types of information, enabling KERL to deliver more precise recommendations.

\subsection{Evaluation on Conversation Task}
\label{conversation}

\begin{table}[]
  \centering
    \caption{Automatic Evaluation on the generation task. The best results are in bold, the second-best results are underlined. ``Improv." denotes the relative improvement compared to the second-best results. ``Dist" refers to Distinct, and ``Ratio" refers to Item Ratio. }
    \label{tab:gen-auto}
    \renewcommand{\arraystretch}{1.3}
    \resizebox{\linewidth}{!}{%
     \begin{tabular}{lcccccccc}
        \toprule
            Dataset                & \multicolumn{4}{c}{ReDial} & \multicolumn{4}{c}{INSPIRED} \\ \midrule
            Model                  & Dist-2 & Dist-3 & Dist-4 & Ratio & Dist-2 & Dist-3 & Dist-4 & Ratio \\ \midrule
            Transformer            & 0.148 & 0.151 & 0.137 & 0.194 & 1.02 & 2.248 & 3.582 & 0.07  \\
            GPT-2                  & 0.354 & 0.486 & 0.452 & 0.353 & 2.347 & 3.691 & 4.568 & 0.214  \\ \midrule
            ReDial                 & 0.225 & 0.236 & 0.228 & 0.158 & 1.347 & 1.521 & 3.445 & 0.05 \\
            KBRD                   & 0.263 & 0.368 & 0.423 & 0.298 & 1.369 & 2.259 & 3.592 & 0.141\\
            KGSF                   & 0.330 & 0.417 & 0.521 & 0.325 & 1.608 & 2.719 & 4.929 & 0.201 \\
            C$^2$-CRS              & \underline{0.631}  & \underline{0.932}  & \underline{0.909} & 0.336 & 2.456 & 4.432 & 5.092 & 0.208\\ 
            UniCRS                 & 0.369  & 0.565  & 0.734 & \underline{0.526} & \underline{3.039} & \underline{4.657} & \underline{5.635} & \underline{0.231} \\ \midrule
            \textbf{KERL}          & $\textbf{0.764}$  & $\textbf{1.430}$  & $\textbf{1.919}$ & $\textbf{0.575}$ & $\textbf{3.109}$ & $\textbf{5.224}$ & $\textbf{6.655}$ & $\textbf{0.300}$ \\ 
            Improv.                & 21.1\% & 53.4\% & 111.1\% & 9.31\% & 2.3\% & 12.2\% & 18.1\% & 29.9\% \\ \bottomrule
    \end{tabular}%
}
\end{table}

In this section, we evaluate the effectiveness of the proposed model for the conversation task and report the evaluation results through automatic and human evaluation metrics in Table~\ref{tab:gen-auto} and Table~\ref{tab:gen-human}, respectively.
\newline

\noindent\textbf{Automatic Evaluation}. We can see that ReDial surpasses Transformer, as ReDial utilizes a pre-trained RNN model to generate more precise representations of past conversations.
KBRD achieves better performance than ReDial by incorporating external KG into the generation module.
KGSF further enhances performance by leveraging cross-attention to integrate aligned entity-oriented and word-oriented KGs into context.
Furthermore, GPT-2 achieves better performance than these three CRS methods, which is likely due to its pre-training on generative tasks with a large-scale corpus, enabling it to quickly adapt to the CRS task and generate diverse responses after fine-tuning. 
UniCRS significantly outperforms these models in terms of item ratio on both datasets by injecting task-specific knowledge into the PLM through semantic fusion and prompt learning. This suggests that UniCRS is more effective at incorporating items from the knowledge graph into the generated responses, which is a crucial aspect of CRSs. In terms of diversity metrics, C$^2$-CRS achieves better performance on the ReDial dataset compared to UniCRS. This could be attributed to C$^2$-CRS's contrastive learning approach that incorporates external information such as reviews and knowledge graphs, enabling it to generate more diverse responses within the constraints of shorter response lengths in ReDial. On the other hand, UniCRS has better performance on the INSPIRED dataset, which has longer average response lengths. The longer responses in INSPIRED provide UniCRS with more opportunities to showcase its ability to generate diverse and informative responses by leveraging its unified knowledge-enhanced prompt framework and pre-trained model.

In comparison to these baselines, KERL shows significant improvements in all evaluation metrics. We leverage textual descriptions of entities and incorporate this knowledge into our knowledge graph, which is then further incorporated into the response generation module. This enables our model to better understand the contextual information related to the entity, resulting in more informative and entity-aware responses. Moreover, we employ a copy mechanism that enhances the diversity and entity awareness of the generated responses by directly copying relevant entity information from the knowledge graph. We demonstrated that our approach can significantly enhance the performance of the conversation task.
\newline

\begin{table}[t]
    \centering
    \caption{Human Evaluation on the generation task.
        The best results are in bold, the second-best results are underlined.
        Gen and CRS are text generation methods and conversational recommendation methods, respectively.}
    \label{tab:gen-human}
    \renewcommand{\arraystretch}{1.3}
    \begin{tabular*}{\linewidth}{c@{\extracolsep{\fill}}lcc}
        \toprule
        \textbf{Type} & \textbf{Model} & \textbf{Fluency} & \textbf{Informativeness}   \\ \midrule
        \multirow{2}{*}{{\textbf{Gen}}}&Transformer        & 0.82    & 0.91          \\
        &GPT-2              & 1.51    & 1.15           \\       \midrule
        \multirow{6}{*}{{\textbf{CRS}}}&ReDial              & 0.95    & 0.87           \\
        &KBRD               & 0.88    & 0.96            \\
        &KGSF               & 1.03    & 1.20             \\
        &C$^2$-CRS          & 1.21   & \underline{1.45}       \\
        &UniCRS             & \underline{1.82}    & 1.15   \\ \cmidrule{2-4}
        &\textbf{KERL}      & \textbf{1.89}    & \textbf{1.70}           \\ \bottomrule
    \end{tabular*}
\end{table}

\noindent\textbf{Human Evaluation}. We conduct human evaluation to measure the quality of the generated responses based on fluency and informativeness.
From Table \ref{tab:gen-human}, we can see that KERL outperforms all baseline models in both metrics, demonstrating its superior ability to understand and incorporate contextual information related to the entity. This superior performance aligns with KERL's high item ratio scores in the automatic evaluation, confirming its effectiveness in generating informative and entity-aware responses by incorporating textual descriptions of entities into the knowledge graph and leveraging cross-attention and the copy mechanism.
In contrast, C$^2$-CRS enhances informativeness through contrastive learning and cross-attention but struggles with fluency due to its non-PLM-based architecture.
In our manual annotation, we observe that UniCRS tends to generate more fluent but template-like responses with less entity-related information.
Considering UniCRS's high item ratio scores in the automatic evaluation, this finding suggests that while UniCRS is capable of incorporating items into its responses, it may not always provide meaningful and contextually relevant information about those items. The comparison between PLM-based and non-PLM-based CRS models highlights the importance of leveraging pre-trained language models for generating fluent and coherent responses.

Overall, the comparison of these models shows the effectiveness of KERL's approach, which strikes an optimal balance between producing coherent, fluent dialogue and embedding rich entity-related information.

\begin{table*}[t]
  \centering
  \caption{Results of the ablation and variation study on the recommendation task. The percentage decrease in performance compared to the original model is shown as superscripts. D and PE refer to entity description and entity positional encoding. KE and CL refer to the knowledge embedding method and contrastive learning.}
  \label{tab:ab}
  \renewcommand{\arraystretch}{1.3}
  \begin{tabular*}{\textwidth}{l@{\extracolsep{\fill}}lllllll}
    \toprule
    \multicolumn{1}{l}{\textbf{Model}} & \textbf{Variant} & \multicolumn{3}{c}{\textbf{Redial}} & \multicolumn{3}{c}{\textbf{INSPIRED}} \\
    \cmidrule(lr){3-5} \cmidrule(lr){6-8}
    & & \textbf{R@1} & \textbf{R@10} & \textbf{R@50} & \textbf{R@1} & \textbf{R@10} & \textbf{R@50} \\ \midrule
    \multirow{5}{*}{KERL}      & \textbf{Original}  & \textbf{0.056} & \textbf{0.217} & \textbf{0.426} & \textbf{0.106} & \textbf{0.281} & \textbf{0.439} \\
    & w/o D   & $0.047^{-0.16}$ & $0.206^{-0.05}$ & $0.405^{-0.05}$ & 0.084$^{-0.21}$ & 0.251$^{-0.11}$ & 0.409$^{-0.07}$ \\
    & w/o PE  & $0.048^{-0.14}$ & $0.207^{-0.04}$ & $0.406^{-0.05}$ & 0.097$^{-0.08}$ & 0.252$^{-0.10}$ & 0.419$^{-0.04}$ \\
    & w/o KE  & $0.050^{-0.10}$ & $0.210^{-0.03}$ & $0.425^{-0.00}$ & 0.105$^{-0.01}$ & 0.264$^{-0.06}$ & 0.419$^{-0.04}$ \\
    & w/o CL  & $0.048^{-0.14}$ & $0.211^{-0.03}$ & $0.421^{-0.01}$ &  0.094$^{-0.11}$ & 0.280$^{-0.00}$ & 0.416$^{-0.05}$ \\ \midrule
    \multicolumn{1}{l}{C$^2$-CRS} & \textbf{Original}  & 0.050 & 0.218 & 0.402 & 0.089 & 0.240 & 0.382 \\
    & w/o CL  & $0.031^{-0.38}$ & $0.149^{-0.31}$ & $0.338^{-0.16}$ & 0.058$^{-0.34}$ & 0.174$^{-0.28}$ & 0.263$^{-0.31}$ \\ \bottomrule
  \end{tabular*}%
\end{table*}

\subsection{Ablation Studies and Analysis}

In this section, we present the ablation study and analysis of our proposed KERL model to verify the effectiveness of each component.
\newline

\subsubsection{Ablation Studies} We evaluate the performance~of different variants of our model by removing one component at a time while keeping the other components unchanged. Specifically, we consider the following variants:
(1) \textbf{KERL~\textit{w/o}~Description}, which removes entity description and uses Xavier initializer \cite{pmlr-v9-glorot10a} to initialize entity embeddings;
(2) \textbf{KERL~\textit{w/o}~PE}, which removes the positional encoding;
(3) \textbf{KERL~\textit{w/o}~KE}, which removes the TransE-based knowledge embedding learning objective;
(4) \textbf{KERL~\textit{w/o}~CL}, which removes the contrastive learning;

As shown in Table \ref{tab:ab}, removing the entity descriptions leads to a significant performance drop across both datasets, with decreases of 16\% and 21\% in Recall@1, respectively. This significant decrease in performance emphasizes the critical role of semantic context provided by these descriptions in enhancing the model's ability to understand and differentiate between entities. Without this semantic layer (i.e. PLM enhanced knowledge embeddings), the model struggles to distinguish between entities with similar structural characteristics but differing contextual meanings. This highlights the importance of integrating descriptive and semantic information with structural graph data for effective CRSs.
Similarly, the absence of positional encoding results in decreased performance, showcasing its effectiveness in capturing the temporal information of entities within a conversation.
Furthermore, omitting the knowledge embedding learning objective results in a decrease in performance, reflecting its importance in understanding entity relationships. The effect is even more significant when CL is removed, resulting in a 14\% drop in Recall@1 for the ReDial dataset and a similar trend observed in the INSPIRED dataset.
This highlights the important role of CL in the model, particularly in its ability to align and synthesize different types of information for effective recommendations. The lack of CL notably impairs the model's ability to integrate diverse data sources, which is crucial for accurately inferring user preferences and interests in dynamic conversational contexts.
Moreover, a comparison between \textbf{KERL~\textit{w/o}~CL} and \textbf{C$^2$-CRS~\textit{w/o}~CL} reveals a much smaller degradation in KERL, illustrating its robustness and adeptness at utilizing text-based entity representations and contextual information.
The resilience of KERL, even without alignment mechanisms like CL, points to its effectiveness in fusing different types of information in CRSs.

\subsubsection{Influence of Different Pooling Methods}
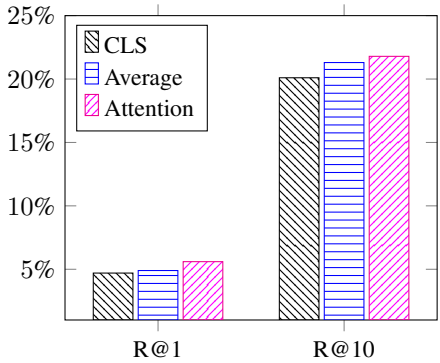
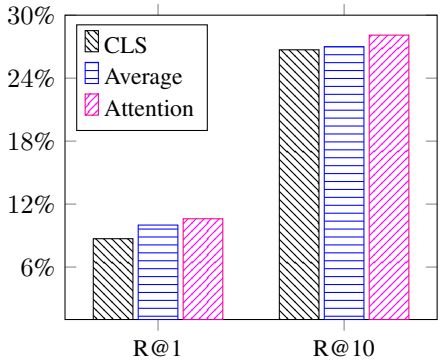
\begin{figure}[t]
  \centering
  \begin{subfigure}[b]{0.45\textwidth}
    \centering
    \begin{tikzpicture}
    \begin{axis}[
        ybar,
        bar width=15,
        symbolic x coords={R@1, R@10, R@50},
        xtick=data,
        xticklabel style={align=center, font=\small},
        enlarge x limits=0.5,
        ymin=0.01,
        ymax=0.25, 
        ytick={0.05, 0.10, 0.15, 0.20, 0.25},
        yticklabel={\pgfmathparse{\tick*100}\pgfmathprintnumber{\pgfmathresult}\%},
        width=0.8\columnwidth,
        legend pos=north west,
        legend image code/.code={
            \draw [/tikz/.cd,bar width=5pt,yshift=-0.2em,bar shift=0pt]
            plot coordinates {(0cm,0.8em)};
          },
        legend style={
            fill=none,
            font=\small,
            draw=black,
          },
        legend cell align={left},
      ]
      \addplot[black, fill=white, pattern=north west lines, pattern color=.,] coordinates {(R@1,0.047) (R@10,0.201)};
      \addlegendentry{CLS}
      \addplot[blue, fill=white, pattern=horizontal lines, pattern color=., ] coordinates {(R@1,0.049) (R@10,0.213)};
      \addlegendentry{Average}
      \addplot[magenta, fill=white, pattern=north east lines, pattern color=.,] coordinates {(R@1,0.056) (R@10,0.2179)};
      \addlegendentry{Attention}
    \end{axis}
    \end{tikzpicture}
    \caption{ReDial}
    \label{fig:redial}
  \end{subfigure}%
  \hfill
  \begin{subfigure}[b]{0.45\textwidth}
    \centering
    \begin{tikzpicture}
     \begin{axis}[
        ybar,
        bar width=15,
        symbolic x coords={R@1, R@10, R@50},
        xtick=data,
        xticklabel style={align=center, font=\small},
        enlarge x limits=0.5,
        ymin=0.01,
        ymax=0.30, 
        ytick={0.06, 0.12, 0.18, 0.24, 0.30},
        yticklabel={\pgfmathparse{\tick*100}\pgfmathprintnumber{\pgfmathresult}\%},
        width=0.8\columnwidth,
        legend pos=north west,
        legend image code/.code={
            \draw [/tikz/.cd,bar width=5pt,yshift=-0.2em,bar shift=0pt]
            plot coordinates {(0cm,0.8em)};
          },
        legend style={
            fill=none,
            font=\small,
            draw=black,
          },
        legend cell align={left},
      ]
      \addplot[black, fill=white, pattern=north west lines, pattern color=.,] coordinates {(R@1,0.087) (R@10,0.267)};
      \addlegendentry{CLS}
      \addplot[blue, fill=white, pattern=horizontal lines, pattern color=., ] coordinates {(R@1,0.100) (R@10,0.270)};
      \addlegendentry{Average}
      \addplot[magenta, fill=white, pattern=north east lines, pattern color=.,] coordinates {(R@1,0.106) (R@10, 0.281)};
      \addlegendentry{Attention}
    \end{axis}
    \end{tikzpicture}
    \caption{INSPIRED}
    \label{fig:inspired}
  \end{subfigure}
  \caption{Comparison of different pooling methods on the ReDial and INSPIRED datasets.}
  \label{fig:recall_pooling}
\end{figure}

We also explore using different pooling methods for learning entity embeddings from entity descriptions using the entity text encoder.
We compare three methods, namely:
(1) \textbf{CLS}, which utilizes the representation of the ``[CLS]" token as the entity embedding. This method is a widely used method for obtaining sentence embeddings;
(2) \textbf{Average}, which computes the average of the hidden states of the entity text encoder;
(3) \textbf{Attention}, which uses an attention network to learn entity embeddings from the hidden states.
As shown in Figure \ref{fig:recall_pooling}, the CLS method yields the worst performance, as it cannot exploit all the output hidden states of the entity text encoder.
The Attention method outperforms the Average method, since the attention network can selectively focus on informative parts of the entity description while disregarding irrelevant information. This results in a more accurate and informative representation of the entity, compared to the Average method, which assigns equal weights to all tokens. Therefore, the attention mechanism allows for a more fine-grained analysis of textual information, enabling the model to capture the semantic meaning of entities more effectively.

\subsubsection{Convergence and Performance}
\begin{figure*}[t] 
  \centering
  \subfloat[Recall@1 for Different Model Variants\label{fig:subfig_a}]{%
    \includegraphics[width=0.31\textwidth]{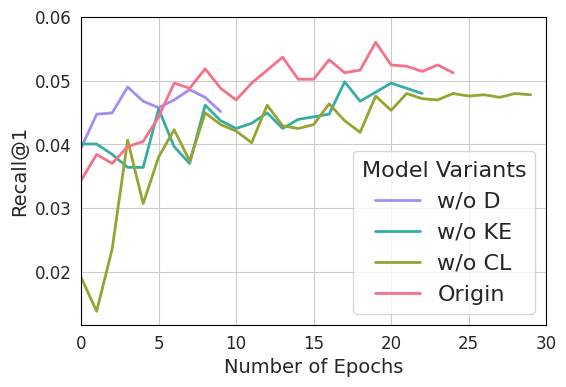}
  }
  \hfill
  \subfloat[Recall@10 for Different Model Variants\label{fig:subfig_b}]{%
    \includegraphics[width=0.31\textwidth]{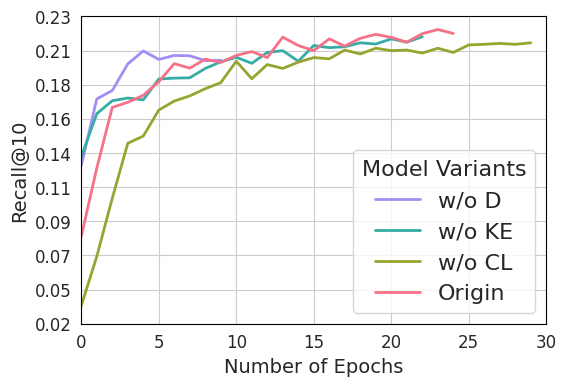}
  }
  \hfill
  \subfloat[Recall@50 for Different Model Variants\label{fig:subfig_c}]{%
    \includegraphics[width=0.31\textwidth]{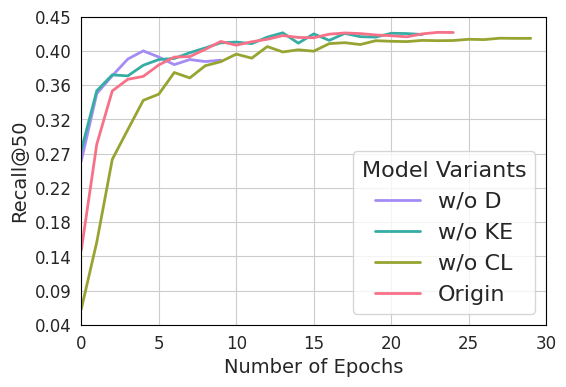}
  }
  \caption{Performance comparison of different KERL variants on recommendation tasks on the ReDial Dataset. D, KE and CL refer to entity description, knowledge embedding method and contrastive learning.}
  \label{fig:recall_convg}
\end{figure*}

In this section, we examine the convergence rate and overall performance of the proposed model and its variants. We incrementally increase the number of epochs during training and report the corresponding test set performance. Training ceases when the model's performance plateaus after three epochs of the evaluation set.
Figure \ref{fig:recall_convg} shows the performance variation as the number of iterations increases. Notably, KERL outperforms its non-contrastive learning counterpart with fewer training iterations. This pattern underscores the efficiency of the CL component in KERL, suggesting its role in contributing to faster and more robust convergence.
Moreover, KERL utilizing solely relation information, without entity descriptions, has superior performance in the initial epochs.

This can be attributed to its effectiveness in capturing and utilizing straightforward, structural patterns inherent in the KG. 
For example, the model can quickly learn to recognize and recommend based on simply yet significant patterns, such as users' preferences for movies by specific directors. This early advantage likely arises because these structural patterns are more readily discernable and simpler to model.
However, when training surpasses a specific epoch threshold, the model overfits the relationship information, leading to a marginal performance improvement. 
This indicates a ceiling effect where relational data alone is no longer sufficient for continued performance gains.
Conversely, integrating both relationship information and entity descriptions initially resulted in worse performance, implying an increase in complexity in handling multiple data modalities, as the model must learn to effectively combine both types of information.
However, as training progresses, this integrative approach begins to yield better results. This improvement over time shows the model's evolving capability to interpret entity characteristics and their interrelationships more effectively.
Ultimately, this comprehensive understanding of entities and their relationships can produce superior performance, as the model fully leverages the richness of the combined data.
\newline

\subsubsection{Influence of Entity Description Length}
To study the impact of entity description length on both recommendation and conversation tasks, we conduct experiments by varying the maximum number of tokens allowed in entity descriptions.
As shown in Table \ref{tab:len}, most evaluation metrics achieved the best results at a maximum length of 40 tokens.
While longer descriptions generally provide more information about the entity and can improve the performance of the CRS,
excessive length may introduce noise and hinder the model's ability to effectively capture the key features of the entity, leading to a reduction in performance.
Furthermore, excessive length can increase computational costs.
Thus, in our approach, if an entity description exceeds 40 tokens, only the initial 40 tokens are utilized, with the rest being disregarded. Despite the potential loss of valuable information through this process, the conducted experiments confirm that this approach achieves a good balance between performance and efficiency.

\begin{table}[]
  \centering
  \caption{The effect of entity description length on the recommendation and conversation tasks for the ReDial dataset. Bold indicates the best results, and the second-best results are underlined. }
  \label{tab:len}
  \renewcommand{\arraystretch}{1.3}
  \begin{tabular*}{\linewidth}{c@{\extracolsep{\fill}}cccccc}
    \toprule
    
    \multirow{2}{*}{\textbf{Length}} & \multicolumn{3}{c}{\textbf{Recommendation}} & \multicolumn{3}{c}{\textbf{Conversation}}        \\ \cmidrule{2-7}
    & \textbf{R@1}   & \textbf{R@10}  & \textbf{R@50}           & \textbf{Dist@2} & \textbf{Dist@3} & \textbf{Dist@4} \\ \midrule
    20 & 0.047 & 0.215 & 0.425          & 0.566  & 1.085  & 1.545  \\
    30 & 0.049 & 0.210 & 0.425          & 0.695  & 1.275  & 1.706  \\
    40 & \textbf{0.056}   & \textbf{0.217}   & $\underline{0.426}$   & \textbf{0.764} & \textbf{1.430} & \textbf{1.919} \\
    50 & $\underline{0.050}$ & $\underline{0.215}$ & \textbf{0.430} & $\underline{0.733}$  & $\underline{1.323}$  & $\underline{1.792}$  \\ \bottomrule
  \end{tabular*}%
\end{table}
\subsection{Case Study}

\begin{table}[]
    \centering
    \caption{Case Study showcasing generated responses from different models on movie item recommendation. Italic blue font indicates the movie item and red font indicates the entity.}
    \label{tab:case-study}
    \begin{tabular}{p{0.2\linewidth}p{0.7\linewidth}}
        \toprule
        \textbf{Recommender}: & What kind of movies do you like?                                                                \\
        \textbf{User}:        & I am looking for a movie recommendation. When I was younger, I really enjoyed
        \textbf{\textit{\textcolor{blue}{A Nightmare on Elm Street (1984)}}}.                                                    \\\midrule
        \textbf{UniCRS}:      & I love that movie!                                                                              \\
        \textbf{C$^2$CRS}:    & Yeah, that was a good one.                                                                      \\
        \textbf{KERL(Ours)}   & I really like that movie!                                                                       \\ \midrule
        \textbf{User}:        & I also enjoyed watching \textbf{\textit{\textcolor{blue}{The Last House on the Left (1972)}}}. I do enjoy some of the newer horror movies that I have seen as well.
  \\ \midrule
        \textbf{UniCRS}:      & I have not seen that one.                                                                       \\
        \textbf{C$^2$CRS}     & Yeah, \textbf{\textit{\textcolor{blue}{Rosemary's Baby (1968)}}} classic classic
        \textbf{\textcolor{red}{slasher}} flick.                                                                                \\
        \textbf{KERL(Ours)}   & \textbf{\textit{\textcolor{blue}{It (2017)}}} is a classic \textbf{\textcolor{red}{horror}}
        movie with a lot of twists.                                                                                             \\ \midrule \midrule
        \textbf{Recommender}: & Hello there. Looking for some good movies?                                                      \\
        \textbf{User}:        & I am looking for a nice movie to watch tonight alone.                                           \\
        \textbf{Recommender}: & Any specific genre?                                                                             \\
        \textbf{User}:        & Maybe a \textbf{\textcolor{red}{mystery}}.                                                      \\ \midrule
        \textbf{UniCRS}:      & \textbf{\textit{\textcolor{blue}{Murder on the Orient Express (2017)}}} is good one.            \\
        \textbf{C$^2$CRS}:    & Have you seen the movie \textbf{\textit{\textcolor{blue}{Hannibal (2001)}}}?                    \\
        \textbf{KERL(Ours)}:  & Okay, great. Have you watched \textbf{\textit{\textcolor{blue}{Seven (1995)}}}? It's a classic. \\ \midrule
        \textbf{User}:        & I also like \textbf{\textcolor{red}{crime}} movies,
        \textbf{\textit{\textcolor{blue}{Hannibal (2001)}}} was great!                                                          \\ \midrule
        \textbf{UniCRS}:      & \textbf{\textit{\textcolor{blue}{The Silence of the Lambs (1991)}}} is a good one too.                 \\
        \textbf{C$^2$CRS}:    & Then \textbf{\textit{\textcolor{blue}{Zodiac (2007)}}} is another classic.                      \\
        \textbf{KERL(Ours)}:  & If you like \textbf{\textcolor{red}{crime thrillers}} then
        \textbf{\textit{\textcolor{blue}{Shutter Island (2010)}}} is a great choice.                                            \\ \bottomrule
    \end{tabular}%
\end{table}

In this section, we present a case study using two conversation examples displayed in Table~\ref{tab:case-study} to demonstrate the quality of generated responses and recommended items by our proposed model, KERL, in comparison to UniCRS and C$^2$CRS.

Our model effectively integrates dialogue context and background knowledge to generate informative responses and appropriate recommendations, thereby enhancing the user experience. For instance, in the first example, when the user expresses their preferences, KERL identifies the user's interest in horror movies and recommends ``It (2017)", a newer film that combines intense horror scenes with violent supernatural elements. This recommendation aligns with the user's favored films, ``A Nightmare on Elm Street (1984)" and ``The Last House on the Left (1972)", both known for similar thematic content. In contrast, C$^2$CRS recommends ``Rosemary's Baby (1968)", a film that, while also a horror, primarily focuses on psychological horror and does not cater to the user's preference for slasher-style horror, and is also older. 
Moreover, in the second example, KERL initially suggests ``Seven (1995)", a highly-regarded mystery movie, in response to the user's request. When the user reveals their appreciation for crime movies and mentions ``Hannibal (2001)", KERL adapts by recommending ``Shutter Island (2010)". This film, a psychological thriller within the crime genre, mirrors the elements appreciated in ``Hannibal (2001)" and reflects KERL's ability to incorporate new information from the conversation and adjust its recommendations accordingly.
Conversely, C$^2$CRS, while providing genre-appropriate recommendations, lacks depth in adapting to the user's psychological preferences. UniCRS, after learning the user's enjoyment of ``Hannibal (2001)", recommends ``The Silence of the Lambs (1991)", which, although relevant, does not expand on the psychological depth or introduce diverse themes.

This case study emphasizes the superiority of our proposed model in comprehending user preferences and providing context-aware recommendations and informative responses. In contrast, C$^2$CRS occasionally faces difficulties in delivering fluent responses and may generate repeated tokens, as seen in the first example, resulting in a degraded user experience. On the other hand, UniCRS tends to generate generic responses with less item-related information. 

\section{Conclusion}\label{sec:conclusion}

In this paper, we presented a novel framework, Knowledge-Enhanced Entity Representation Learning, for conversational recommendation systems.
Specifically, our approach leverages a knowledge graph with entity descriptions to better understand the intrinsic information in each entity, and employs an entity encoder to construct entity embeddings from textual descriptions.
We then used positional encoding to capture the temporal information of entities in the conversation to ensure more accurate summaries of user preferences.
By merging entity information into knowledge-enhanced BART model and copy mechanism, we were able to generate fluent and informative responses.
Moreover, we constructed a high-quality KG, namely WikiMKG, to facilitate our study.

As future work, we intend to focus on improving the interpretability and explainability of generated recommendations. This involves developing techniques to provide transparent reasoning for each recommendation, which is crucial for mitigating potential inaccuracies and biases, thereby ensuring more reliable and user-focused results. 
Subsequently, the exploration will extend to the utilization of pre-trained user profiles for CRSs.
Learning from pre-trained profiles can be more realistic and common in the real world, as users often provide explicit preferences or have a history of interactions with the system that can be leveraged to personalize recommendations.



\bibliographystyle{IEEEtran}
\bibliography{references}







\end{document}